\documentclass[letterpaper, 10 pt, conference]{ieeeconf}  

\IEEEoverridecommandlockouts                              

\usepackage[dvipsnames]{xcolor}
\usepackage{amsmath}
\usepackage{amssymb}
\usepackage{graphicx}
\usepackage[ruled,vlined]{algorithm2e}
\usepackage{hyperref}

\usepackage[per=slash]{siunitx}
\usepackage{comment}
\usepackage[normalem]{ulem}
\usepackage{xspace}
\usepackage{xcolor}

\definecolor{OliveGreen}{RGB}{0,200,25}
\newcommand{\red}[1]{\textcolor{red}{#1}}

\newcommand{\darkgreen}[1]{\textcolor{OliveGreen}{#1}}

\newcommand{\ie}{i.\,e.\ }
\newcommand{\eg}{e.\,g.\ }

\newcommand{\ackInvasic}{The research leading to these results has received funding from the Deutsche Forschungsgemeinschaft (DFG, German Research Foundation) – Project Number 146371743 – TRR 89 Invasive Computing.}

\newcommand{\added}[1]{\darkgreen{#1}}
\newcommand{\replaced}[2]{\red{\ifmmode\text{\sout{\ensuremath{#1}}}\else\sout{#1}\fi}\darkgreen{#2}}
\newcommand{\removed}[1]{\red{\ifmmode\text{\sout{\ensuremath{#1}}}\else\sout{#1}\fi}}

\newcommand{\removedfootnote}[1]{\footnote{\removed{#1}}}
\newcommand{\removedsubsection}[1]{\subsection{\texorpdfstring{\removed{#1}}{#1}}}

\newcommand{\R}{\mathbb{R}}

\title{\LARGE \bf
Graph-based Task-specific Prediction Models for Interactions between Deformable and Rigid Objects
}

\author{Zehang Weng*$^{1}$, Fabian Paus*$^{2}$, 
Anastasiia Varava$^{1}$, Hang Yin$^{1}$, Tamim Asfour$^{2}$ and Danica Kragic$^{1}$
\thanks{*Authors with equal contribution.}
\thanks{$^{1}$The authors are with CAS/RPL, KTH, Royal Institute of Technology, Stocholm,  Sweden. {\tt\small \{zehang,varava,hyin,dani\}@kth.se}}%
\thanks{$^{2}$The authors are with the Institute for Anthropomatics and Robotics, Karlsruhe Institute of Technology, Karlsruhe, Germany. {\tt\small \{paus,asfour\}@kit.edu}}%
}

\begin{document}

\maketitle
\thispagestyle{empty}
\pagestyle{empty}

\begin{abstract}


Capturing scene dynamics and predicting the future scene state is challenging but essential for robotic manipulation tasks, especially when the scene contains both rigid and deformable objects. In this work, we contribute a simulation environment and generate a novel dataset for task-specific manipulation, involving interactions between rigid objects and a deformable bag. The dataset incorporates a rich variety of scenarios including different object sizes, object numbers and manipulation actions.
We approach dynamics learning by proposing an object-centric graph representation and two modules which are Active Prediction Module (APM) and Position Prediction Module (PPM) based on graph neural networks with an encode-process-decode architecture. 
At the inference stage, we build a two-stage model based on the learned modules for single time step prediction. We combine modules with different prediction horizons into a mixed-horizon model which addresses long-term prediction.
In an ablation study, we show the benefits of the two-stage model for single time step prediction and the effectiveness of the mixed-horizon model for long-term prediction tasks.
Supplementary material is available at \url{https://github.com/wengzehang/deformable_rigid_interaction_prediction}
\end{abstract}

\section{Introduction} \label{intro}


Predicting action effects is essential for robotic manipulation. Models capturing task scenes are usually incorporated in predictive control to achieve some specific manipulation goals~\cite{ye2020object} or facilitating sensing in interactive perception. While multiple works address rigid object manipulation, modeling and predicting the scene dynamics of highly-deformable objects such as cloth, which is essential for many real-life tasks~\cite{sanchez2018robotic}, remains challenging. As a potential solution, learning-based modeling~\cite{battaglia2016interaction} accommodates unmodeled effects of physical simulators and is applicable to various task representations. In this paper, we focus on predicting the dynamics of interactions between both rigid and cloth-like objects in a simulated environment. 
Building learning-based predictive models for scenes is challenging for several reasons. First, there is currently no publicly available dataset containing complex interactions with highly deformable objects. Second, 
generalization requires an effective model that captures scenes with internal and external relations of a varying number of scene objects.



\begin{figure}[tb]
    \centering
    \includegraphics[width=1.0\linewidth]{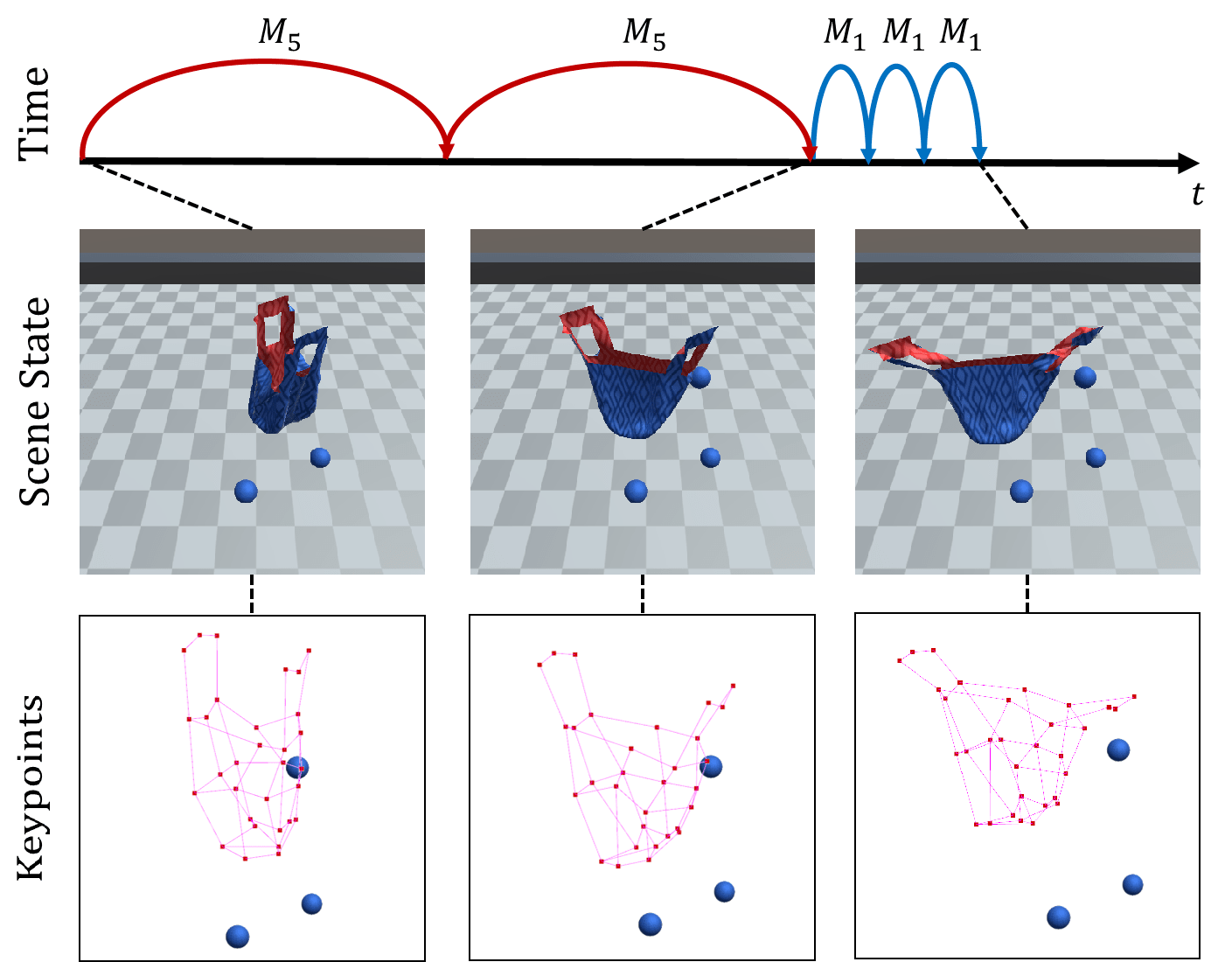}
    \caption{The mixed-horizon model consists of a short-term prediction model $M_1$, which can predict the next time step, and a long-term prediction model $M_5$, which can predict five time steps into the future.
    This figure shows the scene state at different time steps and our sparse keypoint representation of the scene state at these time steps.
    }
    \label{fig:conecpt}
\end{figure}

Recent works typically process simulation data of objects with simple topologies, such as ropes~\cite{seita2020learning}, and scenes with limited objects and interactions, such as cloth dropping~\cite{lin2020softgym}. Various graph-based approaches towards dynamics learning have been proposed~\cite{battaglia2016interaction,watters2017visual,li2020learning}. Typically, these works consider 
low-level physics of particles and limited interactions between them, instead of a task-level representation. 
We contribute a dataset for learning action effects on scenes with both deformable and rigid objects.
To this end, we  build a simulation environment modeling the interaction between several rigid spheres and a deformable bag with handles using Unity and the Obi Cloth extension. 
We collect data for 20 different tasks including four different actions with varying material and environment settings.
Depending on the task, we select a sparse set of keypoints on the deformable object's surface and represent the scene state as a fully-connected graph. We learn task-specific dynamics models based on two separated graph modules for single time step predictions. Based on these dynamics models, we propose a mixed-horizon model for predicting the action effects over multiple time steps, which combines single time step models with different prediction horizons (see Fig.~\ref{fig:conecpt} for an example).

The proposed model can be used for various deformable object manipulation tasks, such as arranging objects, opening the bag, putting the objects into the bag, and deforming the bag by pushing another object towards it.
We evaluate the proposed method on all 20 tasks in the dataset and show the advantages of our model in single time step predictions as well as in long-horizon prediction tasks.
A video illustrating the dataset and the proposed method is available\footnote{\url{https://youtu.be/a4ILwCmai9k}}.

The main contributions of the paper are: 
\begin{itemize}

    \item We build a novel publicly available dataset\footnote{\url{https://github.com/wengzehang/deformable_rigid_interaction_prediction/blob/main/docs/dataset.md}}
    for task-specific action effect prediction by leveraging a deformable simulation engine. The scene contains interactions between one cloth-like deformable object and multiple rigid objects.
    
    \item We propose a method for predicting complex interactions between deformable and rigid objects by representing the scenes as graphs and building a two-stage prediction model, which first classifies which parts of the scene move at all and then applies position updates selectively in a second stage. 
    By combining two-stage models with different prediction horizons, our method outperforms baseline approaches with only one prediction horizon.

\end{itemize}

The rest of the paper is organized as follows: in Section~\ref{relatedwork}, we provide an overview of the related work. In Section \ref{method_modeling}, we describe the scene modeling and dataset generation.  Section \ref{method_learning} presents our graph-based approach to dynamics prediction. In Section \ref{evaluation}, we evaluate our approach.
The paper concludes in Section~\ref{sec:conclusion} with a discussion of the results.

\section{Related Work}\label{relatedwork}

\subsection{Deformable Object Dynamics Modeling}
In deformable object manipulation, there are two types of methods for predicting complex object dynamics as a result of an action. The first one is based on analytical modeling, Hou et al. review different traditional methods for cloth dynamics modeling~\cite{hou2019review}. One popular approach to accurate modeling is applying the finite element method (FEM), 
which mostly applies to fabrics of simple shapes for real-time simulations~\cite{sanchez2018robotic}.
Another approach is to construct a particle-based simulation system based on the measured mechanical properties (friction, mass, elasticity, bending, etc.) by standardized measurement systems like “Kawabata Evaluation System" (KES) and "Fabric Assurance by Simple Testing System" (FAST)~\cite{luible2008simulation}. However, these methods are computationally expensive, especially when the geometric and topological structure is complex.

The second type of methods relies on learning the dynamics from data. Here, the  dynamics are captured without explicitly measuring and memorizing the object properties. Recently, many research works addressed rope dynamics modeling. Battaglia et al. investigate the power of graph-based interaction networks and learn the dynamics of a simulated rope environment~\cite{battaglia2016interaction}. Watters et al. use a front-end network to encode the visual input as latent representations and builds a dynamics estimator based on interaction network structure~\cite{watters2017visual}. Yan et al. takes images as input and uses a neural network to encode the rope state as a set of connected nodes, and apply a bi-directional LSTM to capture the dynamics based on the node representations~\cite{yan2020self}. For 2D cloth-like objects, a physically-based simulator and fully-connected networks are combined to perform the simulation~\cite{oh2018hierarchical} and~\cite{lee2019efficient} for coarse and fine levels respectively. PlaNet~\cite{hafner2019learning} encodes the images by an Autoencoder as a latent code and predicts the future latent representation based on GRU structure. PlaNet is evaluated on SoftGym~\cite{lin2020softgym} cloth manipulation dataset but fail to produce accurate estimation.

All these methods have their limitations in complex scenarios. First, most of these works are devoted solely to modeling 1D linear objects like ropes or cables. When 2D cloth-like objects are considered, the authors implicitly assume that their topology is simple. Second, the considered actions are typically restricted to picking and placing. In this work, we study the scene dynamics considering more complicated deformable object and a rich set of actions. 

\subsection{Predicting Action Effects for Rigid Objects}

Predicting action effects for rigid objects has been studied more extensively than for deformable objects.
Early works predicted planar motions of a single object which was represented with a binary segmentation mask~\cite{omrvcen2009autonomous,kopicki2011learning}.
More recent works can handle a fixed number of objects indirectly by predicting the perceived image after action execution~\cite{byravan2017se3, finn2016unsupervised}.
Some image-based methods are limited to a fixed number of objects due to their use of a constant amount of image masks.
Other methods do not use masks but rather predict the complete image after action execution~\cite{agrawal2016learning,eitel2020learning}.
These methods are able to learn dynamics models for non-planar object interactions.
For instance, Zeng et al. leverage depth information by including a height map in the scene representation~\cite{zeng2018learning}.

A new trend in action effect prediction employs graph neural networks to learn system dynamics with the ability to generalize to scenarios with a different number of objects~\cite{battaglia2016interaction}.
Graph neural networks have been also used to predict the motion of stacked block towers~\cite{janner2019reasoning}, the effects of pushing into a scene of rigid objects~\cite{paus2020predicting}, and of interacting with a connected set of rigid objects~\cite{tekden2020belief}.



\subsection{Physical Simulations for Data Collection}
Collecting data for training a predictive model on real hardware is expensive. Simulated environments provide a cheap alternative, but are challenging to set up for highly-deformable objects. In this work, we rely on a simulated environment created using the Unity engine. 


Advanced wrap-up software is used for generating cloth-like object animation, such as Blender, Maya, Unreal Engine, and Unity. The first two are based on the Bullet engine~\cite{coumans2013bullet} and the other two are based on the PhysX/FleX engine~\cite{macklin2014unified}. All these engines use particle-based solvers.~\cite{erez2015simulation} compares these popular physics engines.

The heavy simulation time cost and the unreality of synthetic data increase the difficulty of cloth-like object research. There are very few works about cloth-like object manipulation. Regarding the deformable object benchmark, SoftGym~\cite{lin2020softgym} creates a set of simulation environments with 1D cables and 2D fabrics, based on the FleX engine, and tests the standard reinforcement learning algorithms on their released benchmark. However,  SoftGym does not provide an interface for 3D deformable cloth-like object data collection. Based on the Bullet engine, DeformableRavens~\cite{seita2020learning} creates 12 scenarios with 1D cable, 2D fabric, and 3D bag manipulations, and proposes a goal-conditioned variant of \added{a} Transporter Network for action recommendations on different tasks. However, the used bag is constructed from simple templates without any hole structures on the body. The objects in the scenes are manipulated with "pick" and "place" actions, while the interactions between the bag and rigid objects are induced by gravity. In our work, we design the bag templates with handles to include multi-hole structures using Blender, and build the simulation environment in Unity based on Obi Cloth extension~\cite{obicloth}. As mentioned in Section \ref{intro}, we pre-program a moving sphere or the handle action trajectories to generate rich interactions between different objects in the scene.


\section{Task Description and Dataset Generation} \label{method_modeling}


We generate a novel dataset for task-specific action effect prediction on scenes containing interactions between a deformable bag and a set of rigid objects.

\subsection{Task Description}

\begin{figure}[!htb]
    \centering
    \includegraphics[width=0.45\linewidth]{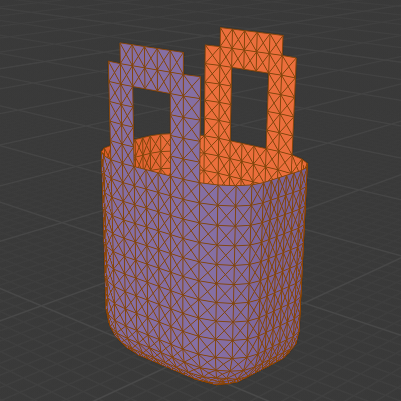}
    \includegraphics[width=0.45\linewidth]{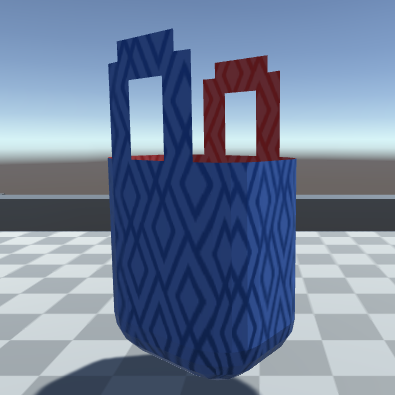}
    \caption{The deformable bag in its initial pose. The first figure is the bag template in Blender. The second figure is the bag in the Unity environment.}
    \label{fig:bagblender}
\end{figure}

We consider tasks like opening a bag, pushing an object into a bag and moving a handle of the bag along a specific trajectory.
A task consists of a parameterized action,
the objects in the scene, and further task parameters like how stiff the bag is (\emph{Bag Stiffness}), whether the bag is empty or a rigid object is inside (\emph{Bag Content}), and whether the handles are fixed in place, loose or moved along a trajectory (\emph{Handle State}).
Each scene contains a deformable bag, some number of rigid spheres, and a table. The bag can interact with rigid objects and the table.
For the deformable bag, we model the mesh in Blender as shown in Fig.~\ref{fig:bagblender}. Compared to the cloth-like objects in previous works, our model has a more complicated hole structure. The whole mesh consists of 1277 particles and 4326 edges.


For the actions, we consider pushing an object towards the deformable bag, moving a handle of the bag along a circular trajectory, opening the bag, and lifting the bag.
The handle motions are achieved by grasping the top part of a handle and moving it along a trajectory.

\begin{figure}[!htb]
    \centering
    \includegraphics[width=0.9\linewidth]{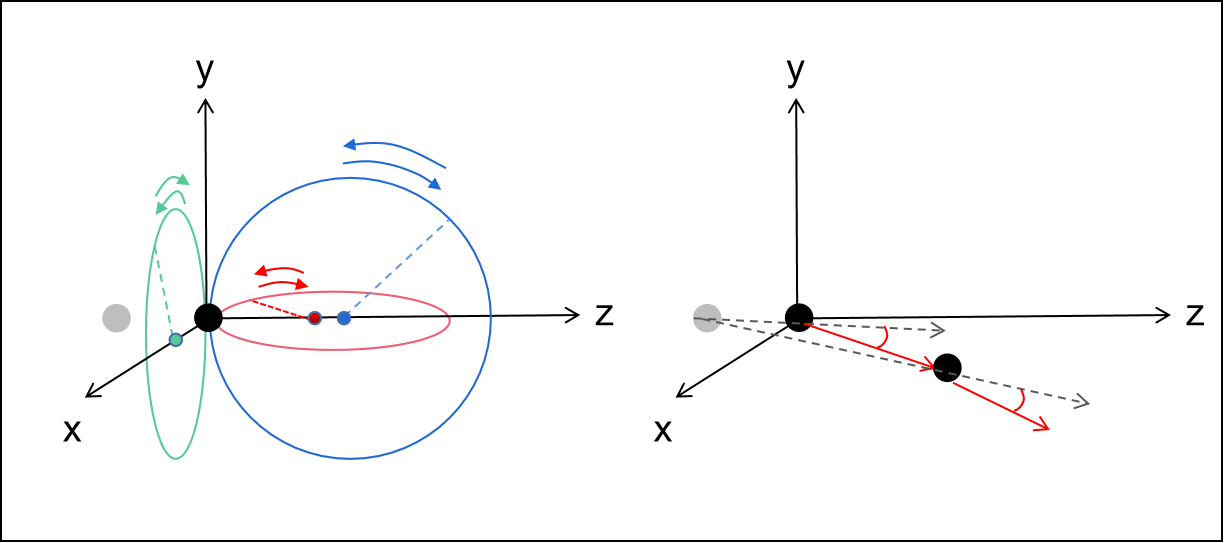}
    \caption{Handle actions. The black point is the manipulated handle and the gray point is the non-targeted handle. The left figure shows examples of circular handle movement in three different coordinate planes. The right figure shows examples of opening actions.}
    \label{fig:handle_movement}
\end{figure}

\begin{itemize}
    \item \emph{Pushing an Object towards the Bag:} 
    We sample a position to create a sphere with a random radius around an existing object.
    A push trajectory is generated by sampling a planar moving direction pointing to either the bag or one of the other rigid objects.
    By applying this strategy, we ensure that most of the actions lead to meaningful object interactions.
    
    
    \item \emph{Handle Motion along Circular Trajectory:} We move one handle along a circular trajectory as shown in Fig.~\ref{fig:handle_movement}. 
    The trajectory is placed in one of the coordinate planes. 
    The radius and direction are randomized.
    
    
    
    \item \emph{Opening the Bag} 
    As shown in Fig.~\ref{fig:handle_movement}, we move one handle away from the other fixed handle in order to stretch the bag horizontally. Before performing the manipulation, we randomly choose a small horizontal deflection angle. During the manipulation, we calculate a base directional vector depending on the handle position differences and rotate it by using the deflection angle to construct the final moving vector. 
    
    
    
    \item \emph{Lifting the Bag:} 
    Before performing this action, the bag is dropped on the table. 
    Then, one handle performs an upward motion, which lifts the bag from the table. 
    The other handle is left loose.
    
\end{itemize}

\subsection{Dataset Generation}

\begin{table*}[]
\centering
\caption{Task Parameters for Data Generation}
\label{tab:SceneTable}
\begin{tabular}{|c|c|c|c|c|c|}
\hline
\textbf{Bag Stiffness} & \textbf{Bag Content} & \textbf{Left Handle State} & \textbf{Right Handle State} & \textbf{Controlled Object} & \textbf{Action}                                                  \\ \hline
Soft/Stiff          & Object Inside             & Fixed         & Fixed                                                     & Sphere                                                                   & Pushing an Object \\ \hline
Soft/Stiff          & Empty                     & Fixed         & Fixed                                                     & Sphere                                                                   & Pushing an Object \\ \hline
Soft/Stiff          & Object Inside             & Moving        & Fixed                                                     & Left Hand                                                                   & Circular Handle Motion         \\ \hline
Soft/Stiff          & Empty                     & Moving        & Fixed                                                     & Left Hand                                                                      & Circular Handle Motion         \\ \hline
Soft/Stiff          & Object Inside             & Moving        & Released                                                  & Left Hand                                                                      & Circular Handle Motion        \\ \hline
Soft/Stiff          & Empty                     & Moving        & Released                                                  & Left Hand                                                                      & Circular Handle Motion         \\ \hline
Soft/Stiff          & Object Inside             & Moving        & Fixed                                                     & Left Hand                                                                      & Opening the Bag             \\ \hline
Soft/Stiff          & Empty                     & Moving        & Fixed                                                     & Left Hand                                                                      & Opening the Bag              \\ \hline
Soft/Stiff          & Object Inside             & Moving        & Released                                                     & Left Hand                                                                      & Lifting the Bag             \\ \hline
Soft/Stiff          & Empty                     & Moving        & Released                                                     & Left Hand                                                                      & Lifting the Bag             \\ \hline
\end{tabular}
\end{table*}

Our simulation environment is based on Unity and the Obi Cloth~\cite{obicloth} extension.
The Obi physics solver is optimized for real-time cloth simulation and supports particle-level manipulation, rich types of interactions, and editable physical constraints (e.g., distance constraints, bending constraints, and aerodynamics). 

The simulation includes a deformable bag, a table, and multiple rigid spheres with random radii for each task.
Further task parameters are generated as follows.
By adjusting the bending constraints in the solver, we vary the stiffness of the bag material (\emph{Bag Stiffness}).
We either initialize the bag in an empty state or with a rigid sphere inside (\emph{Bag Conent}).
The left and right bag handles are either left loose or grasped (\emph{Handle State}).
If a handle is grasped, it either is fixed in place or moves along a given trajectory (opening, lifting, or circular).

During action execution, we record the complete scene state $10$ times per second.
For every recorded time step, the scene state consists of the vertex positions of the deformable bag, the positions and radii of all rigid objects including the pushed sphere, and the grasped vertices on the left and right handle.
Our goal is to learn task-specific models, therefore our dataset is grouped by task.
For each task, we simulate 1,000 trajectories, which results in 60,000 recorded time steps.
The simulated task data is split into training (80\%), validation (10\%), and test set (10\%).
We vary actions and task parameters according to Table~\ref{tab:SceneTable} to create data for 20 different tasks. 
For each row, we collect data for both bag stiffness values (soft and stiff).
Fig.~\ref{fig:scenarios} shows examples for simulated tasks.

\newcommand{\scenewidth}{0.118\linewidth}

\begin{figure*}[!htp]
    \centering
        \includegraphics[width=\scenewidth]{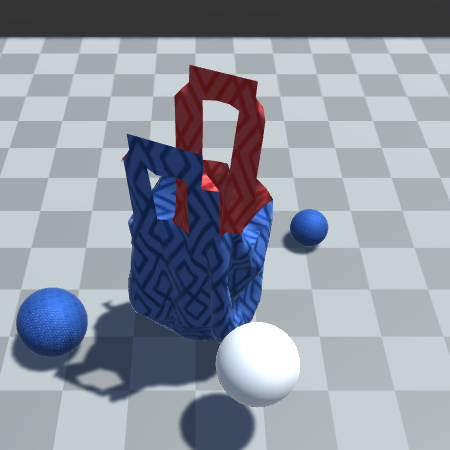}
        \includegraphics[width=\scenewidth]{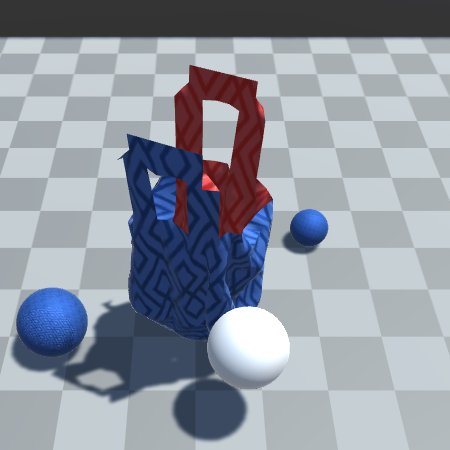}
        \includegraphics[width=\scenewidth]{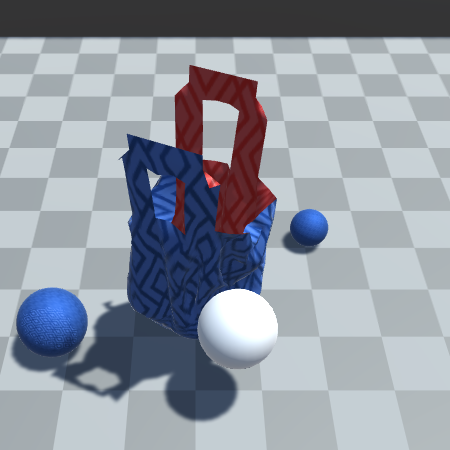}
        \includegraphics[width=\scenewidth]{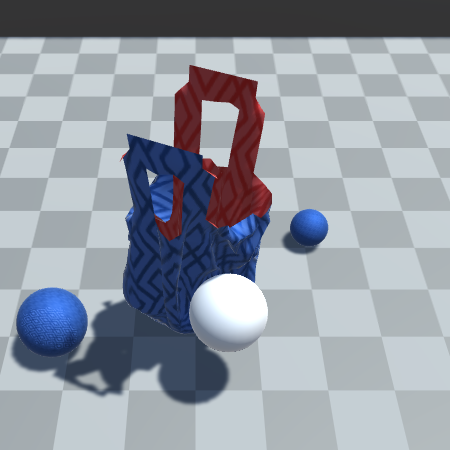}
        \includegraphics[width=\scenewidth]{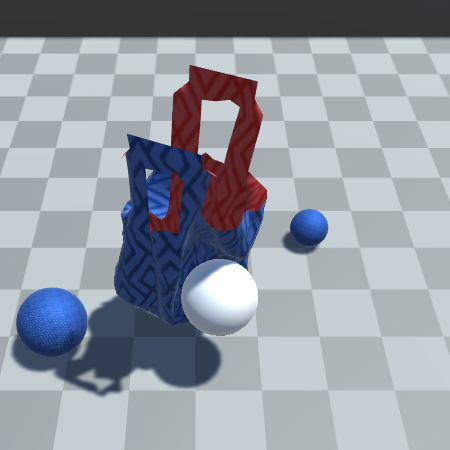}
        \includegraphics[width=\scenewidth]{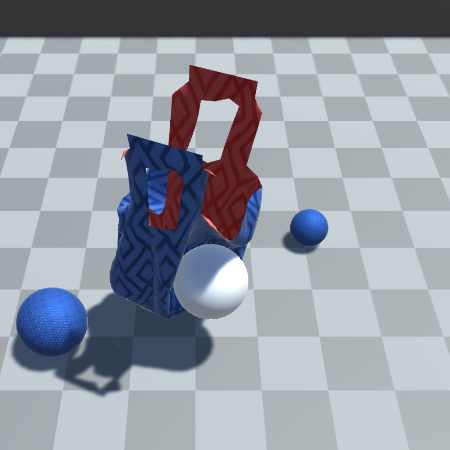}
        \includegraphics[width=\scenewidth]{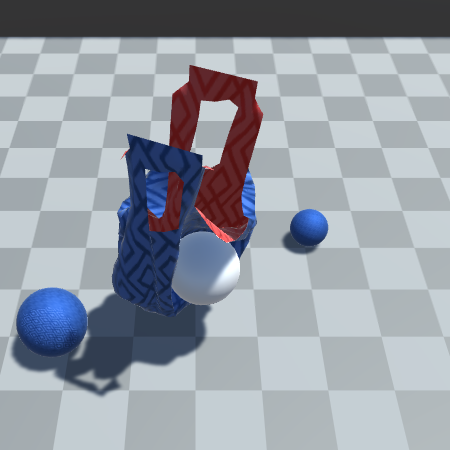}
        \includegraphics[width=\scenewidth]{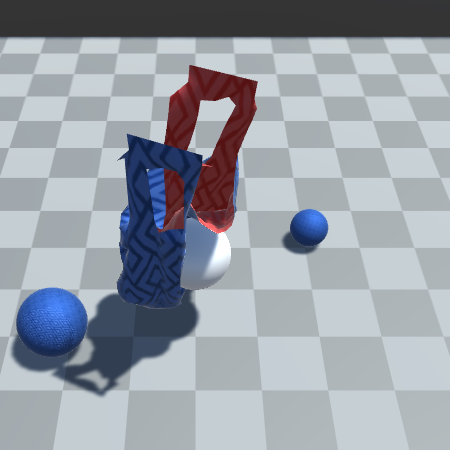}
        
        \includegraphics[width=\scenewidth]{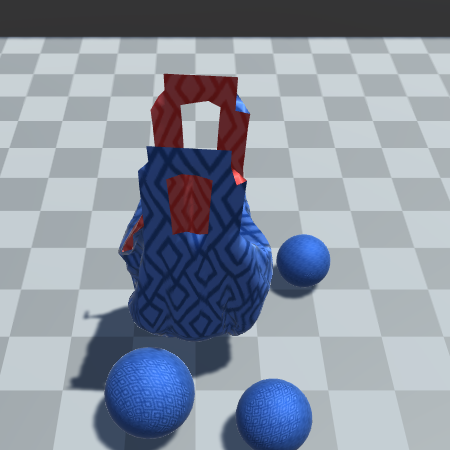}
        \includegraphics[width=\scenewidth]{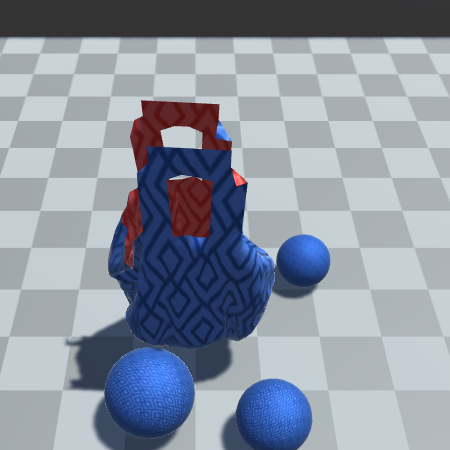}
        \includegraphics[width=\scenewidth]{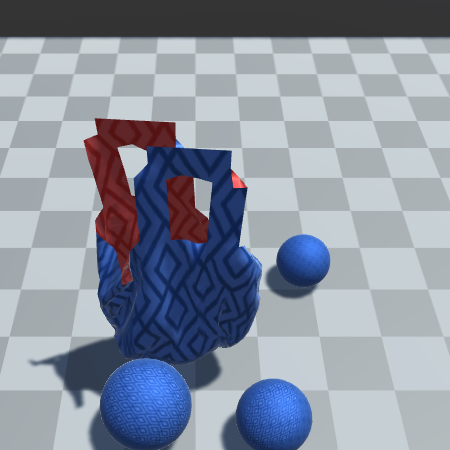}
        \includegraphics[width=\scenewidth]{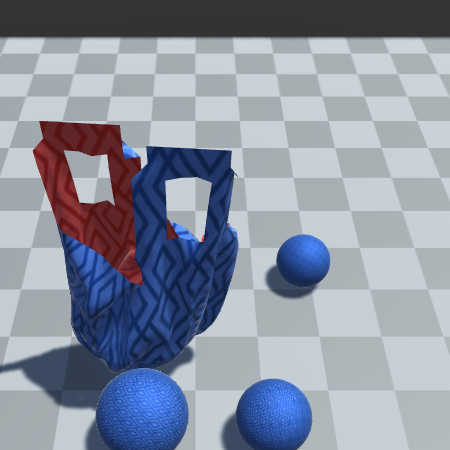}
        \includegraphics[width=\scenewidth]{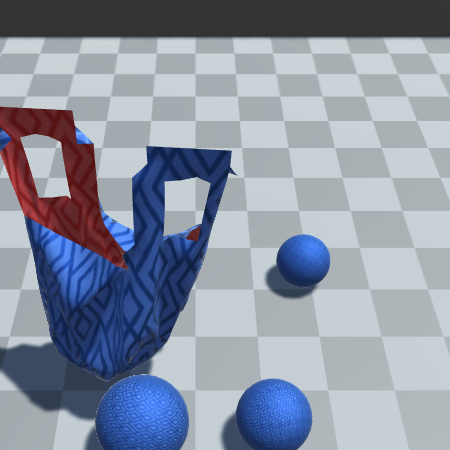}
        \includegraphics[width=\scenewidth]{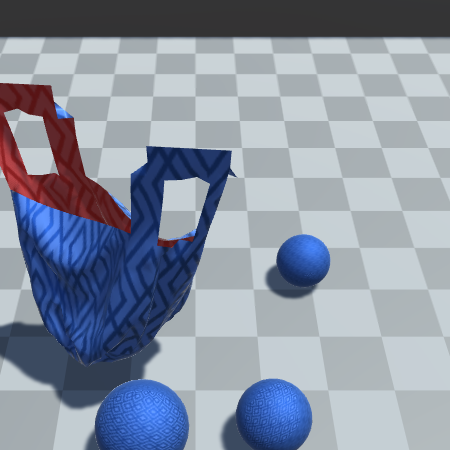}
        \includegraphics[width=\scenewidth]{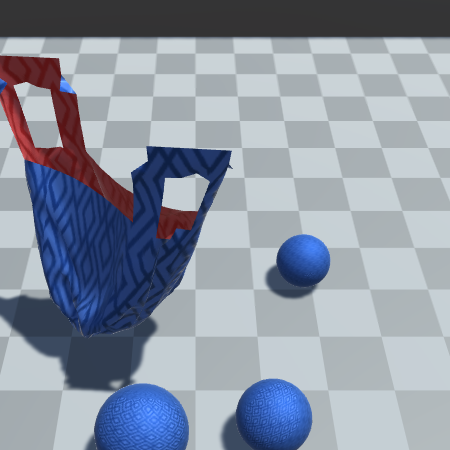}
        \includegraphics[width=\scenewidth]{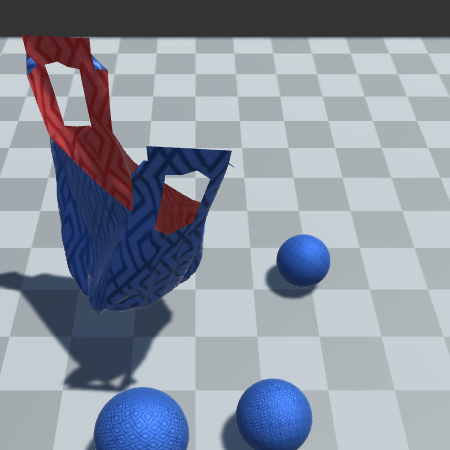}
        
        \includegraphics[width=\scenewidth]{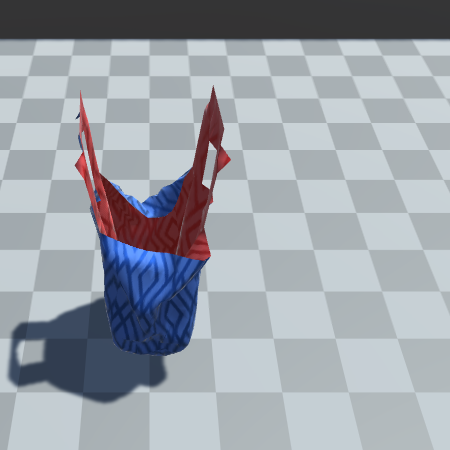}
        \includegraphics[width=\scenewidth]{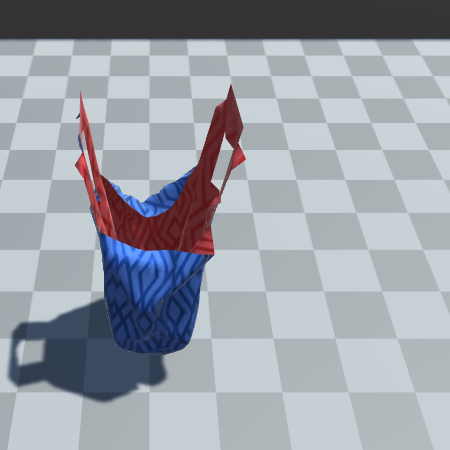}
        \includegraphics[width=\scenewidth]{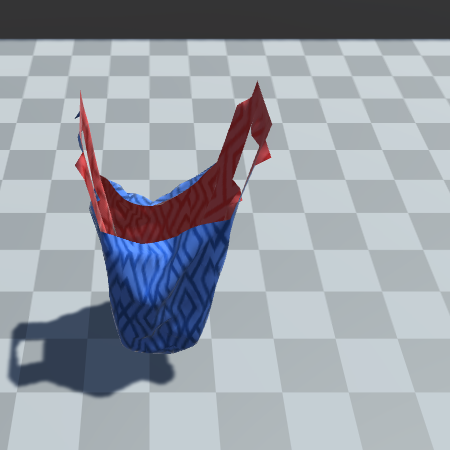}
        \includegraphics[width=\scenewidth]{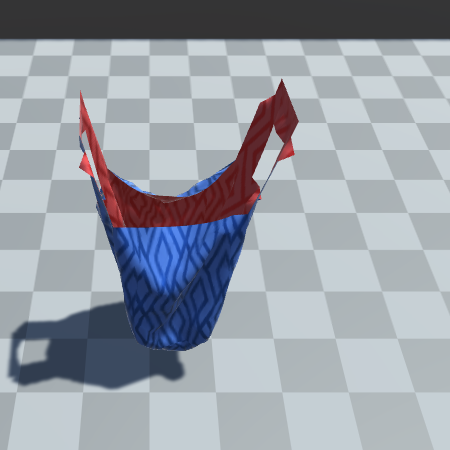}
        \includegraphics[width=\scenewidth]{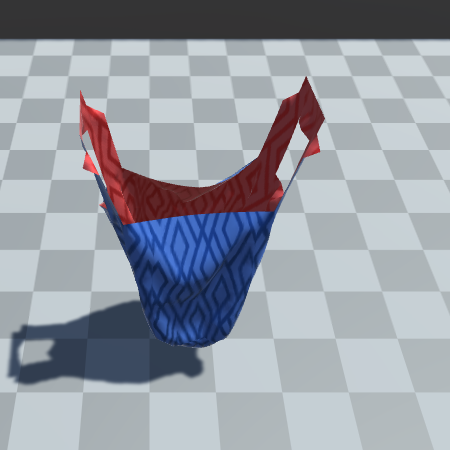}
        \includegraphics[width=\scenewidth]{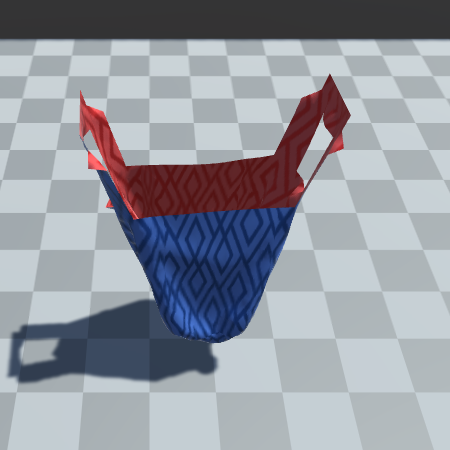}
        \includegraphics[width=\scenewidth]{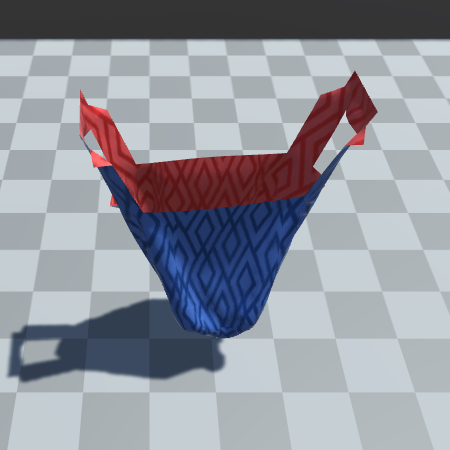}
        \includegraphics[width=\scenewidth]{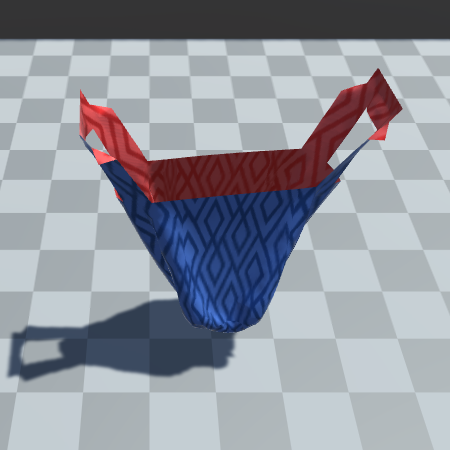}
        
        \includegraphics[width=\scenewidth]{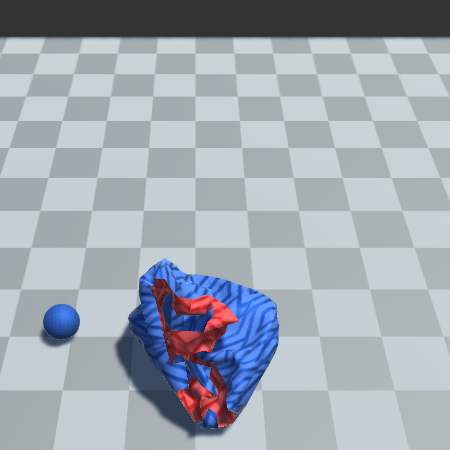}
        \includegraphics[width=\scenewidth]{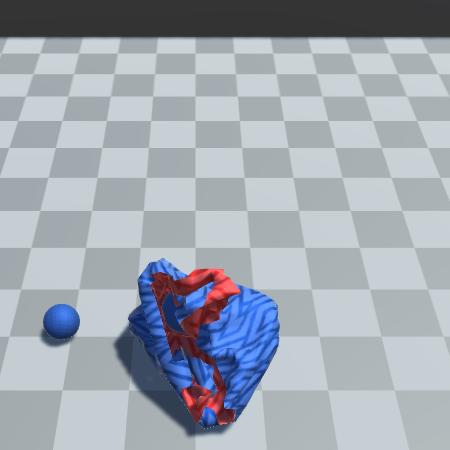}
        \includegraphics[width=\scenewidth]{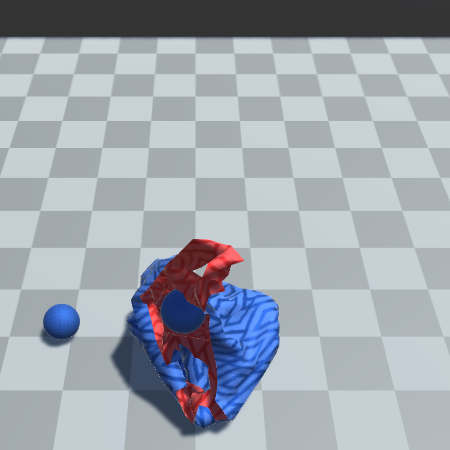}
        \includegraphics[width=\scenewidth]{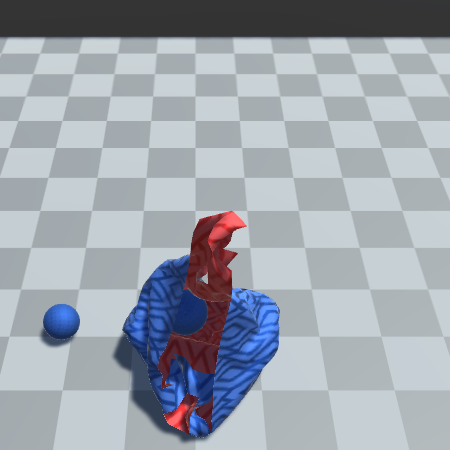}
        \includegraphics[width=\scenewidth]{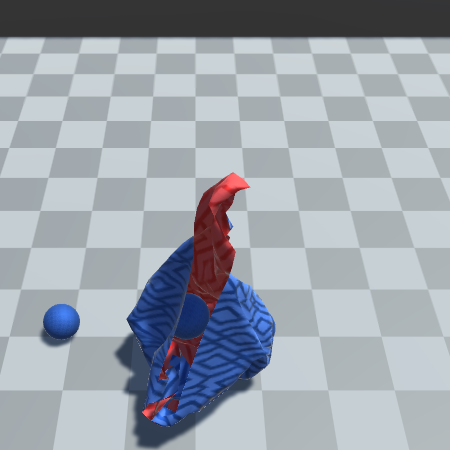}
        \includegraphics[width=\scenewidth]{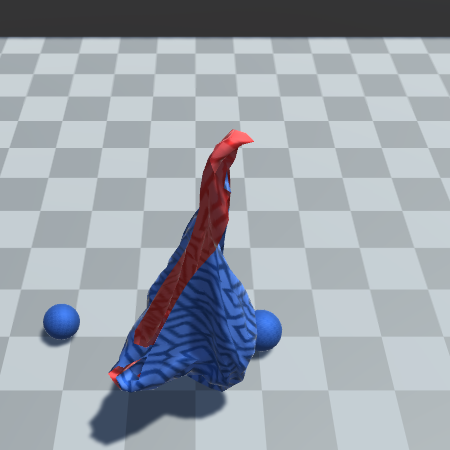}
        \includegraphics[width=\scenewidth]{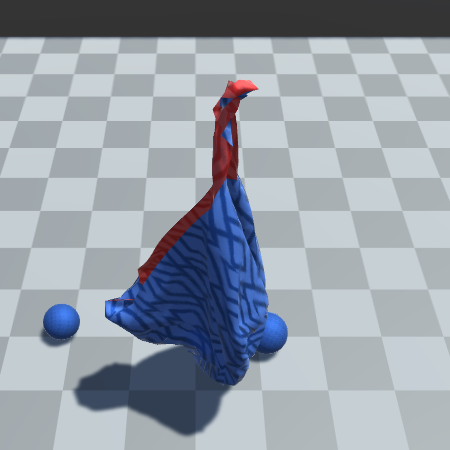}
        \includegraphics[width=\scenewidth]{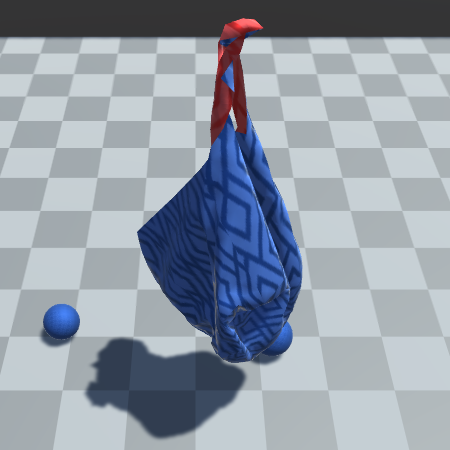}
        
    \caption{Example trajectories of four actions in the dataset.
    Each row contains different time steps of an action.
    From top to bottom: Pushing an Object towards the Bag, Handle Motion along Circular Trajectory, Opening the Bag, and Lifting the Bag.}
    \label{fig:scenarios}
\end{figure*}

\section{Dynamics Learning and Prediction} \label{method_learning}

Based on the generated dataset, we want to learn task-specific prediction models for the scene dynamics.
Given a scene as a set of rigid and deformable objects $O_{t}$, the goal is to learn a dynamics model $M$ to predict the future scene $O_{t+1}$ after performing action $a_{t}$ at time step $t$. 

\begin{equation*}
    \begin{aligned}
        O_{t+1} =  M(O_{t}, a_{t})
    \end{aligned}
    \label{problemdef}
\end{equation*}

The set of rigid objects consists of a variable number of spheres whose state can be represented by their position and radius.
The state of the deformable bag consists of the position and connectivity of all vertices.
The action $a_t$ is parameterized by the start and end position as well as the radius $(\mathbf{p}_{start}, \mathbf{p}_{end}, r_a)$ of the manipulated target,
which can either a rigid object or one of the bag's handles. 

We define a graph representation that captures the state of the rigid objects and approximates the state of the deformable bag using a set of sparse keypoints.
Using this representation, we formulate a two-stage graph learning problem to facilitate fixed time step predictions.
Then, we combine multiple prediction models with different time step horizons to enable predictions of up to 60 time steps into the future.

\subsection{Graph Representation}

\begin{figure}[htb]
    \centering
    \includegraphics[width=1.0\linewidth]{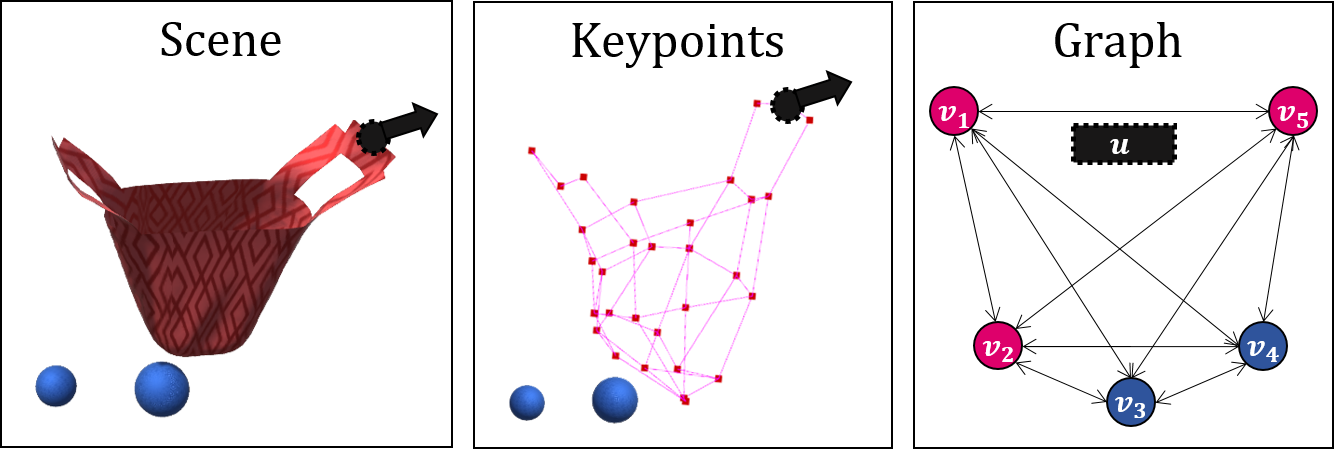}
    \caption{We transform a scene consisting of deformable and rigid objects into a sparse keypoint representation. Based on the keypoints, we build a fully connected graph, whose vertices represent keypoints and whose edges encode the connectivity between them.
    The motion of the handle along the black arrow is encoded in the global graph feature $\mathbf{u}$.
    }
    \label{fig:method:graph-representation}
\end{figure}

\begin{figure*}[tb]
    \centering
    \includegraphics[width=1.0\linewidth]{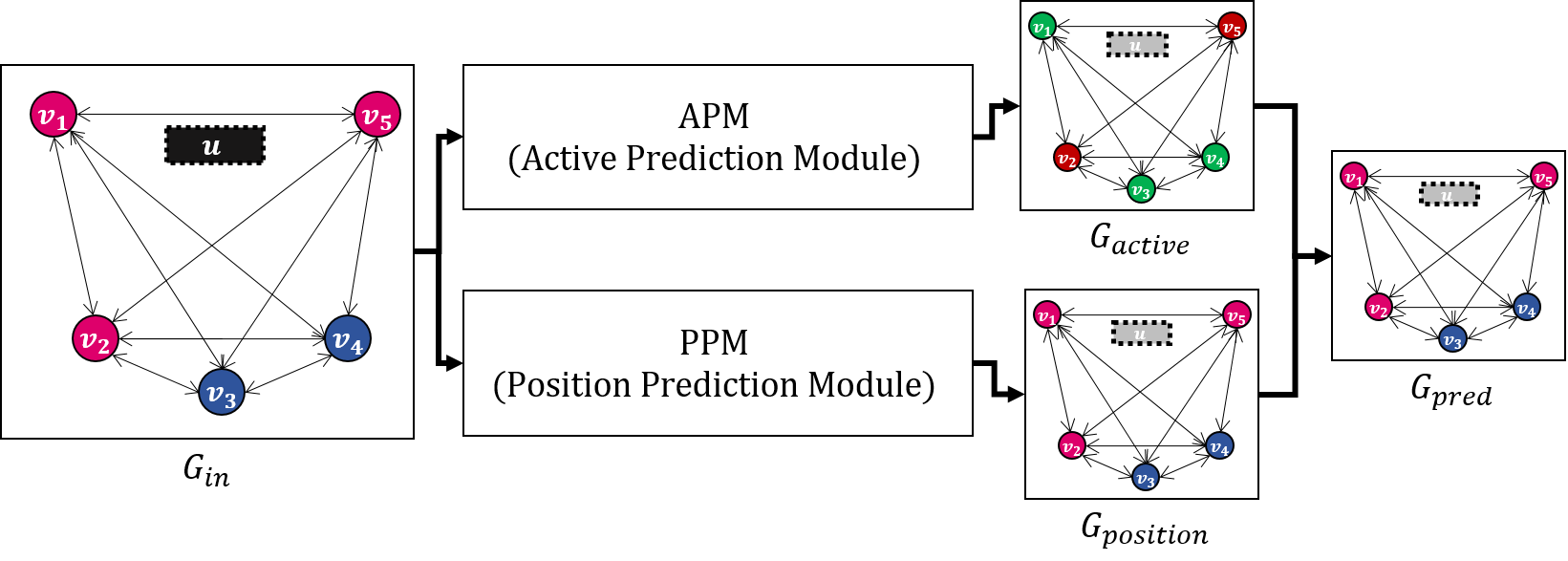}
    \caption{The two-stage model takes as input the scene state as a graph $G_in$ at a certain time step.
    This graph is fed into both the APM and PPM.
    The APM classifies which vertices are active, \ie will move in the next time step.
    In the graph $G_{active}$, the green vertices have been classified as active and the red ones as inactive.
    The PPM predicts the positions of vertices in the next time step as a graph $G_{position}$.
    In a final step, only the position updates, whose corresponding vertices have been classified as active in $G_{active}$, are applied to the prediction result $G_{pred}$.
    }
    \label{fig:method:two-stage}
\end{figure*}

We want to represent the state of the scene objects $O_t$ and the action $a_t$ at time step $t$ as a graph $G_t = (V, E, \mathbf{u})$
with vertices $V$, edges $E$, and a global feature vector $\mathbf{u}$.
The set of vertices $V$ encodes position information about the rigid and deformable objects in the scene (see Fig.~\ref{fig:method:graph-representation}).
We use a vertex feature vector $\mathbf{v}=(\mathbf{t}, r, f) \in \mathbb{R}^5$, which encodes position $\mathbf{t}\in \mathbb{R}^3$, radius $r \in \mathbb{R}$, and a one-hot feature indicating whether the vertex is fixed in place $f \in \{0, 1\}$.
Each rigid object becomes a vertex with the feature vector  $\mathbf{v}=(\mathbf{t}, r, 1) \in \mathbb{R}^5$.
For the deformable bag, we use a sparse keypoint representation, where task-relevant vertices are chosen from the bag's mesh.
Each keypoint becomes a vertex with a feature vector $\mathbf{v}=(\mathbf{t}, 10^{-5}, f)$, where the radius $r$ is set to a small constant value and $f$ indicates whether it can freely move ($f=0$) or is grasped, \ie fixed in place ($f=1$).
Since the choice of a global coordinate system is arbitrary, we transform the positions to an action-local coordinate system, whose origin is the starting position $\mathbf{a}_{start}$ of the manipulated object. 





The edges $E$ build a fully connected bidirectional graph between the vertices $V$.
We use an edge feature vector \mbox{$\mathbf{e}=(\mathbf{d}, f) \in \R^{4}$} consisting of the pairwise position differences $\mathbf{d} \in \R^{3}$ between vertices and the physical connection flag $c \in \{0,1\}$. 
The edges connecting neighboring vertices from the deformable bag have their physical connection flag set to $c=1$. 
All other edges have no direct physical connection ($c=0$).
The global feature vector \mbox{$\mathbf{u}=(\mathbf{p}_{end} - \mathbf{p}_{start}, r_a)\in \R^{4}$} encodes the position change of the manipulated target and the radius of the manipulated object $r_a$.


\subsection{Two-stage Graph Prediction Model}


The goal of the two-stage graph prediction model is, given the current scene state $G_t$, to predict the scene graph $G_{t+h}$ after $h$ time steps where $h$ is constant.
In this work, we focus on single time step predictions ($h=1$) and longer time steps ($h=5$).
For each prediction horizon $h$, we learn a dynamics model which consists of two separate modules: Active Prediction Module (APM) and Position Prediction Module (PPM). APM is a binary classifier predicting whether rigid objects or parts of the deformable bag will move in the next time step.
The classification is done for every vertex in the scene state $G_t$.
The ground-truth active labels are generated during training based on the position difference between the time steps $t$ and $t+h$. 
PPM is a regression module that directly predicts the next scene state, \ie the expected positions of all vertices at time step $t+1$. Both APM and PPM are implemented as graph neural networks with an encode-process-decode architecture.


APM outputs a binary classification mask through a final softmax activation layer for the vertex features.
The classification stage uses cross-entropy loss where $N$ denotes the number of vertices in the scene graph, $y_{i}^{gt} \in \{0, 1\}$ is the ground-truth active flag and $y_{i}^{pred} \in [0, 1]$ is the predicted flag. The active flag is set to be $1$ when the position difference is above a pre-set threshold. 

\begin{equation*}
    \begin{aligned}
        L^{Classification} =  \frac{1}{N} \sum_{i=1}^{N}  CrossEntropy(y_{i}^{gt}, y_{i}^{pred})
    \end{aligned}
\end{equation*}

PPM is a regression model for the scene graph after action execution using a final linear activation layer for the vertex features. 
The regression stage uses a mean square error loss between the ground-truth positions.

\begin{equation*}
    \begin{aligned}
        L^{Regression} = \frac{1}{N} \sum_{i=1}^{N} (\mathbf{t}_{i}^{gt} - \mathbf{t}_{i}^{pred})^2
    \end{aligned}
\end{equation*}

We train both models separately on the tasks in the generated dataset.
By only applying the regression update to those vertices which have been classified as active, we prevent spurious motion of vertices that are not involved in the interaction between objects in the current time step (see Fig.~\ref{fig:method:two-stage}).
Under a fixed time step $h$, we call this combination the \emph{two-stage} model (APM+PPM), whereas the regression stage alone is called \emph{one-stage} model (PPM):

\begin{eqnarray*}
M_{h}^{one-stage}(G_{i}) &=& M_{h}^{PPM}(G_{i}) \\
M_{h}^{two-stage}(G_{i}) &=& M_{h}^{APM}(G_{i}) \odot M_{h}^{PPM}(G_{i})
\end{eqnarray*}
Here the operator $\odot$ only applies the position updates from the PPM if the vertices have been classified as active in the APM.

\subsection{Long Horizon Prediction Model}
\label{sec:long-horizon-prediction}


The graph prediction models only predict the scene for a fixed prediction horizon $h$.  The longer horizon model $M_{5}$ is trained with a prediction horizon $h=5$, and the single time step model $M_{1}$ is trained with a horizon $h=1$. 
By chaining these models recursively together, we can make predictions for any time step $t$.

If we only use the single time step model $M_1$, we can predict the scene state $G_t$ after $t$ time steps given the initial scene state $G_0$:
\begin{equation*}
    G_t = \underbrace{(M_1 \circ M_1 \circ \dots \circ M_1)}_{t \;\text{times}} (G_0)
\end{equation*}


In this approach, we can either use the \emph{one-stage} or the \emph{two-stage} model.
However, this causes the prediction error to accumulate fast.
We can alleviate this problem, by also incorporating the longer horizon model $M_5$.
First, we run $M_5$ recursively for $\lfloor t / 5  \rfloor$ steps.
Then, $M_1$ is run for the remaining time steps $t \mod 5$:
\begin{equation*}
    G_t = 
    \underbrace{(M_1 \circ M_1 \circ \dots \circ M_1)}_{(t\mod 5) \;\text{times}}
    \circ
    \underbrace{(M_5 \circ M_5 \circ \dots \circ M_5)}_{\lfloor t / 5  \rfloor \;\text{times}} 
    (G_0)
\end{equation*}
We call this combination of a multi-step prediction and a single-step prediction the \emph{mixed-horizon} prediction model (see Fig.~\ref{fig:conecpt} for an example).

\section{Evaluation} 
\label{evaluation}

\begin{figure}[tb]
    \centering
        \includegraphics[width=1.0\linewidth]{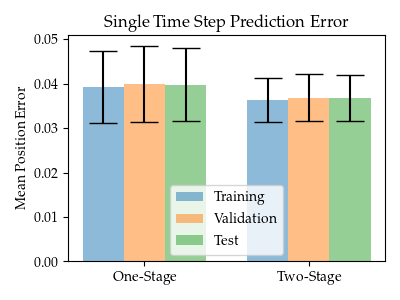}
    \caption{Single time step prediction errors over all tasks for training, validation, and test set.
    The mean position error is shown as the bar height and the whiskers show the standard deviation over all tasks.}
    \label{fig:eval:single-time-step}
\end{figure}

\begin{figure}[tb]
    \centering
        \includegraphics[width=1.0\linewidth]{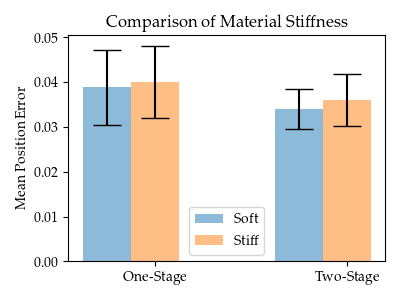}
    \caption{Single time step prediction errors over all tasks grouped by material stiffness.
    The mean position error is shown as the bar height and the whiskers show the standard deviation over all tasks.}
    \label{fig:eval:material-stiffness}
\end{figure}

\begin{figure*}
    \centering
        \includegraphics[width=0.48\linewidth]{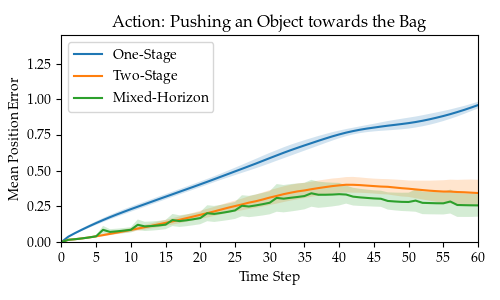}
        \includegraphics[width=0.48\linewidth]{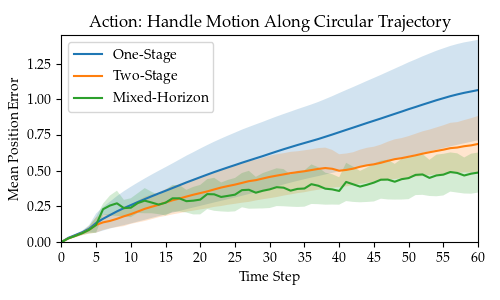}
        \includegraphics[width=0.48\linewidth]{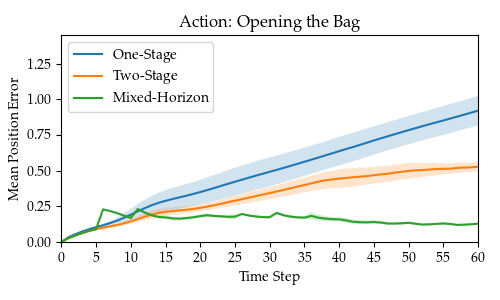}
        \includegraphics[width=0.48\linewidth]{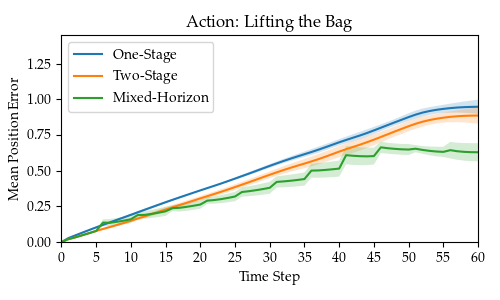}
    \caption{Long horizon prediction errors per action for the \emph{one-stage}, \emph{two-stage}, and \emph{mixed-horizon} models.
    The solid lines show the mean position error while the colored area around the line indicate the standard deviation.}
    \label{fig:eval:long-horizon-per-action}
\end{figure*}

In the evaluation, we want to investigate the benefits of our proposed method by answering the following questions:
\begin{enumerate}
    \item Does the inclusion of the APM (Active Prediction Module) in the \emph{two-stage}  model improve the prediction results over the \emph{one-stage}  model with the PPM (Position Prediction Module) alone?
    
    \item How does the material stiffness of the deformable bag influence the prediction accuracy?

    \item Does the \emph{mixed-horizon} model improve long-term prediction results compared to an iterative application of one-stage or two-stage models?
\end{enumerate}

To answer the first question, we evaluate the single time step prediction performance of the proposed \emph{two-stage} model compared to the \emph{one-stage} model.
Fig.~\ref{fig:eval:single-time-step} shows that the \emph{two-stage} model decreases the mean position errors while also lowering the inter-task variance.
This shows, that the APM improves single time step predictions when compared to the PPM alone.

To address the second question, we compare the single time step prediction results for soft bag material with results for stiff bag material.
Fig.~\ref{fig:eval:material-stiffness} shows the mean position error and the standard deviation for both materials.
As can be seen, the tasks with soft bag material have a smaller prediction error.
However, the difference is lower than the inter-task variance, indicating that our method is able to handle tasks independent of material stiffness.

For the third question, we compare the long horizon prediction results for the recursive \emph{one-stage}, \emph{two-stage} and \emph{mixed-horizon} models on the test set.
We initialize each model with the scene state $G_0$ at time step $t=0$ and apply the prediction in an iterative way as described in section~\ref{sec:long-horizon-prediction}.
Since long horizon prediction performance varies between actions, Fig.~\ref{fig:eval:long-horizon-per-action} shows the mean position errors and standard deviation for the four actions \emph{Pushing an Object towards the Bag}, \emph{Handle Motion along Circular Trajectory}, \emph{Opening the Bag}, and \emph{Lifting the Bag}.
We can see that the \emph{two-stage} model outperforms the \emph{one-stage} model consistently, independent of the action.
The difference between the models in the lifting action is quite small, since the almost all parts of the bag move during this action.
Therefore, the first movement classification stage is not as helpful as in the other actions.
Furthermore, the \emph{mixed-horizon} model outperforms the \emph{two-stage} model for longer term predictions, while sometimes producing worse results for short term predictions.
Depending on the action, the \emph{mixed-horizon} model produces much better predictions then the \emph{two-stage} model (e.g. opening the bag), while for others the improvement is marginal (\eg pushing an object).
Overall, the \emph{mixed-horizon} model is better suited for predictions over a longer time periods than the \emph{one-stage} and \emph{two-stage} models.


\section{Conclusion}
\label{sec:conclusion}

Predicting the dynamics of the scene is important for robotic manipulation, and is difficult in the presence of highly-deformable objects. One big challenge is data collection. In this work, we present a novel dataset for action effect prediction on scenes containing both rigid and cloth-like deformable objects. 
Another challenge is building a predictive model capable of generalizing to different numbers of objects in the scene. We define a graph representation for the scene state, where the vertices are keypoints representing objects and their parts. Our predictive model can generalize to different numbers of vertices in the graph, allowing us to consider different sets of objects. We propose two modules to capture the dynamics based on the graph networks. We propose a mix-horizon model on top of the learned modules to predict the future scene state and show the effectiveness of our methods in different tasks.

In future work, we will investigate meta-learning techniques to accelerate the learning of prediction models for new tasks and reduce the required amount of training data.
Furthermore, we want to study the transfer of prediction knowledge between similar but different tasks, \eg lifting a shirt instead of a bag.

%



%



\section*{ACKNOWLEDGMENT}

\ackInvasic


\IEEEtriggeratref{14}
\bibliographystyle{IEEEtran}
\bibliography{ref}

\end{document}